\documentclass[wcp]{jmlr}

\usepackage{longtable}% for long tables
\usepackage{booktabs}
\usepackage{algorithm}
\usepackage{bbm}
\usepackage{xcolor}
\usepackage{multirow}
\usepackage{amsfonts}
\usepackage{amsmath,amssymb,amsfonts}
\usepackage{algorithmic}
\usepackage{graphicx}
\usepackage{textcomp}
\usepackage{xcolor}
\usepackage{multirow}
\usepackage{makecell}
\usepackage{graphicx}
\usepackage{amsmath}
\usepackage{amssymb}
\usepackage{booktabs}
\usepackage{multicol}
\usepackage{multirow}
\usepackage{bm}
\makeatletter
\let\Ginclude@graphics\@org@Ginclude@graphics 
\makeatother

\title[Dynamic Spectrum Mixer for Visual Recognition]{Dynamic Spectrum Mixer for Visual Recognition}

\author{\Name{Zhiqiang Hu} \Email{zhiqianghu2021@gmail.com}\\
	 \Name{Tao Yu} \Email{yutao@mobile.ee.titech.ac.jp}\\
	}

\begin{document}
	
	\maketitle
	%%%
	%%%%%%%%% ABSTRACT
	\begin{abstract}
		Recently, MLP-based vision backbones have achieved promising performance in several visual recognition tasks. However, the existing MLP-based methods directly aggregate tokens with static weights, leaving the adaptability to different images untouched. Moreover, Recent research demonstrates that MLP-Transformer is great at creating long-range dependencies but ineffective at catching high frequencies that primarily transmit local information, which prevents it from applying to the downstream dense prediction tasks, such as semantic segmentation. To address these challenges, we propose a content-adaptive yet computationally efficient structure, dubbed Dynamic Spectrum Mixer (DSM). The DSM represents token interactions in the frequency domain by employing the Discrete Cosine Transform, which can learn long-term spatial dependencies with log-linear complexity. Furthermore, 
		A dynamic spectrum weight generation layer is proposed as the spectrum bands selector, which could emphasize the informative frequency bands while diminishing others. To this end, the technique can efficiently learn detailed features from visual input that contains both high- and low-frequency information. 
		Extensive experiments show that DSM is a powerful and adaptable backbone for a range of visual recognition tasks. Particularly, DSM outperforms previous transformer-based and MLP-based models, on image classification, object detection, and semantic segmentation tasks, such as 83.8 \% top-1 accuracy on ImageNet, and 49.9 \% mIoU on ADE20K.
		
	\end{abstract}
	\begin{figure*}[tbp]
		\centering
		\includegraphics[width=1\linewidth]{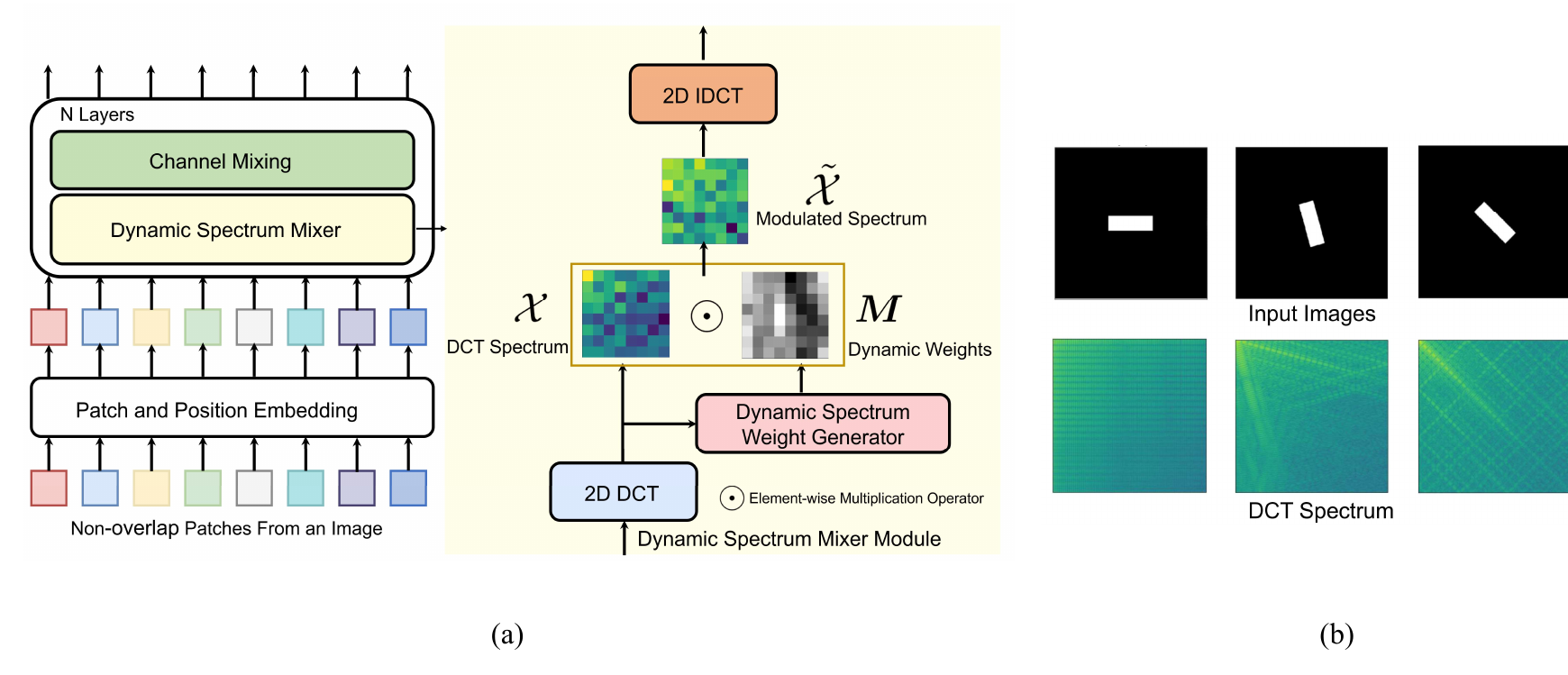}
%		\vspace{5}
		\caption{(a) The architecture of the proposed DSM. The proposed DSM mix the tokens in the frequency domain and empowers them with adaptability to the contents of tokens. (b) The input images and corresponding DCT spectrum. We can see that even rotating the image could cause a dramatic change in the frequency domain. The lighter the color, the larger the DCT spectrum weight }
		%\label{fig:intro}
	\end{figure*}
	%%%%%%%%% BODY TEXT
	\section{Introduction}
	\label{sec:intro}
	
	Convolutional Neural Networks (CNNs) based backbones \cite{he2016deep} have dominated computer vision over the last decade. Recently, inspired by Transformer \cite{vaswani2017attention}, which achieves superior performance in natural language processing, many researchers propose to transfer the transformers into visual recognition tasks e.g., ViT \cite{dosovitskiy2020image}, DeiT \cite{touvron2021training}, and also attained state-of-the-art performance.
	\par More recently, to involve less inductive bias, the MLP-based token Mixer \cite{tolstikhin2021mlp, touvron2022resmlp}, has taken a step forward to entirely abandoning the self-attention layer from the transformer, only employ the channel projections and spatial projections with static parameterization to mix the tokens. Despite the MLP-mixer has achieved promising results in several vision tasks \cite{chen2021cyclemlp, hou2022vision, tolstikhin2021mlp, yu2021s}, there still exists intractable challenges that prevent the model from becoming a universal vision backbone:
	\par (1)These MLP-based approaches utilize fixed static matrices for patch communications or by manually-designed rules in a deterministic manner \cite{chen2021cyclemlp}, which is input agnostic, restrict the adaptability to the contents to be fused, and lead to degradation of representability of learned feature.
	\par (2) MLP-based ViT and its variations are excellent in capturing low-frequency visual data \cite{park2022vision}, mostly global forms and structures of a scene or object, but they are less effective at learning high-frequency visual data, primarily local edges and textures. This is intuitively explained by the fact that the primary operation utilized in MLP to exchange information across non-overlap patch tokens, is a global operation and is far better equipped to capture global information (low frequencies) in the data than local information (high frequencies).
	\par  It has been demonstrated that the Fourier transforms might replace the multi-headed attention layers in transformers and produce equivalent performance. The FNet \cite{eckstein2022fnet}, GFN \cite{rao2021global}, and AFNO \cite{guibas2021adaptive}, which mix the tokens by utilizing Fast Fourier Transform (FFT) in the frequency domain and achieve remarkable accuracy in visual recognition tasks. However, the GFN uses static global filters to exploit the long-term interaction information of spectrum tokens. The global filter is unchanged for different input images. We argue that, the image-agnostic global filter is not the optimal choice. We can see from the toy experiment in Fig. 1 (b), even rotating the image could cause a dramatic change in the frequency domain. Furthermore, these Fourier Transform-based methods treat all the spectrum equally, which is inefficient at capturing high frequencies that primarily carry local information, which prevents it from being used to the downstream dense prediction tasks.
	\par In this paper, we attempt to address these issues mentioned above by seeking a way to project the spatial domain information into image-resolution agnostic transformed space, e.g., frequency domain. Specifically, we propose a dynamic frequency aggregation architecture with DCT and dynamically aggregating them concerning their contents. Each output from DCT has a component of each of the input tokens. Thus the proposed method can facilitate the information interaction across all tokens. In comparison to the DFT, the DCT is a real-valued transform that also breaks down a given signal or picture into its component frequency components. Therefore, in terms of computational cost, DCT is more suited for deep neural networks.
	\par Furthermore, to adaptively amplifies the useful frequency bands while downplaying others, we design a dynamic spectrum band attention module.
	Specifically, we propose simply down-sample the frequency feature to a pre-defined size for the images with variable resolutions and then calculating the attention weights of each spectrum band by using two layers of MLPs. We empirically find that frequency domain feature dimensions can be reduced significantly without sacrificing performance. The dimensions can even be shirked to a tiny number in our experiments. This finding enables us to employ MLPs to explore the rich relationships among spectrum tokens in a computationally efficient fashion. Overall, our contributions can be summarized as follows:
	\begin{itemize}
		\item We propose an MLP-like dynamic spectrum bands mixer, which could mix the tokens in the frequency domain and be empowered with adaptability to the contents of tokens. 
		\item We developed an efficient spectrum band intersection module, which could capture the long-term dependency in a fixed-size down-sampled frequency domain without losing the universal approximating power. We also propose a Dynamic Spectrum Weight Generator (DSWG), which enhances the performance of ViT models by adaptively reweighting the high-frequency and low-frequency components with respect to the input image.  The DSWG can learn representations of the global structure while filtering out frequency space data that is unrelated to the structure.
		\item We achieve 83.8\% top-1 accuracy on ImageNet with 10.1G FLOPs, which significantly surpasses several MPL-like competitors. We also achieve consistent performance in other downstream tasks including object detection and semantic segmentation. 
	\end{itemize}
	%-------------------------------------------------------------------------
	\section{Related Works}
	\subsection{Vision Transformer}
	Vision Transformer (ViT) \cite{dosovitskiy2020image} is the pioneering work that crops the entire image into 16 × 16 patches and treats each patch as a token as the input for the transformer. By utilizing an extremely largescale dataset, JFT-300M \cite{sun2017revisiting}, Vit has achieved promising performance comparable to CNN-based backbones. DeiT \cite{touvron2021training} further employs distillation and data augmentation techniques to improve the performance of ViT, without using an ultra-large-scale training dataset. The concurrent work, PVT \cite{wang2021pyramid}, and PiT \cite{heo2021rethinking} exploit a feature pyramids architecture for vision transformers, which gradually down sampling the spatial dimension, making them more flexible for downstream dense prediction tasks. Han et al. propose TNT \cite{han2021transformer} to model the attention information for the local patches for achieving better performance. To further enhance transferability, Swin Transformer, and Swin transformer v2 \cite{liu2021swin,liu2022swin} propose a new local attention paradigm that employs patch-level multi-head attention equipped with a hierarchical fusion design. Following this way, Shuffle Swin Transformer \cite{huang2021shuffle} proposes shuffle multi-headed attention to reparametrize spatial connection between windows to improve the representation ability. Recently, DynamicViT \cite{rao2021dynamicvit} embraced the dynamic token sparing module with a vision-transformer and significantly increased the efficiency. 
	Vision Transformer also shows its superiority on various downstream dense prediction tasks, e.g., object detection \cite{carion2020end}, segmentation \cite{cheng2022masked}.
	\subsection{MLP-based Architectures}
	\par  Global token-mixing MLPs: The pioneering work MLP-Mixer \cite{tolstikhin2021mlp} employs two Fully-Connected (FC) layers to fulfill the goal of communicating between tokens and achieving results that are comparable with ViT. gMLP \cite{liu2021pay} utilizes a gating operator to enhance the long-term interactions between different spatial locations. Recently, the Wave-MLP \cite{tang2022image} manually represents each token with two parts, amplitude, and phase, followed by a dynamic fusion operation. FNet \cite{eckstein2022fnet} first introduces Fourier Transform as the mixing mechanism and achieves excellent performance. A similar approach, Global Filter Network \cite{rao2021global} also projects each token into the frequency domain by utilizing 2D FFT and then mixes the tokens in the frequency domain via a static global filter. However, the input-agonic globe filter hampers the representability of the model. In contrast, we propose to mix the spectrum tokens dynamically. 
	\par Local token-mixing MLPs: To enhance the efficiency of MLP-based token mixer, local token-mixing MLPs aim to invoke token mixing at the local region.
	AS-MLP \cite{lian2021mlp}, $S^{2}$-MLP and $S^{2}$-MLPv2 \cite{yu2022s2,yu2021s} perform spatial shift operations along the different axis to aggregate information from nearby tokens. CycleMLP \cite{chen2021cyclemlp} and Morphmlp \cite{zhang2021morphmlp} propose mixing the local spatial information by employing hand-craft local kernels in a deterministic way. Hire-MLP \cite{guo2022hire} suggests hierarchically mixing the tokens to further enhance the representing ability. For the hybrid approach, ConvMLP \cite{li2021convmlp} proposes a co-design of the convolution layer and MLP for the downstream dense prediction tasks. 
	\subsection{Frequency Domain Learning}
	CNN-empowered Frequency domain learning has been successfully applied in multiple vision tasks, including Low-level vision such as JPEG image compression \cite{gueguen2018faster}, image super-resolution \cite{magid2021dynamic}, etc. Enhance the feature representations with global receptive fields \cite{chi2020fast}. Frequency domain attention is introduced by FcaNet \cite{qin2021fcanet}, which enhances the representability of ResNet \cite{he2016deep} on the image classification task. Specifically, the work proposed by Xu et al. in \cite{xu2020learning} finds that the down-sampling in the frequency domain can better preserve image information than spatially resizing the images. This motivated us to derive frequency token interaction information in the down-sampled frequency domain.
	\section{Approach}
	In this section, we present the proposed  Dynamic Spectrum Mixer (DSM) in detail.
	After introducing the overall architecture briefly, we present the spectrum bands interaction module, which represents each token from the perspective of the DCT spectrum and
	reweights them in an input-adaptive manner. Finally, we introduce the architecture variants of the proposed DSM.
	\subsection{Preliminaries: Discrete Cosine Transform}
	DCT has been verified as a powerful tool in the image processing area, such as JPEG image compression, due to the properties of concentrating the energy of the image in several coefficients and decorrelating the coefficients. \cite{gueguen2018faster}. 
	Mathematically, the two-dimensional (2D) DCT is formatted as follows:
	\begin{equation}
		B_{h, w}^{i, j}=\cos \left(\frac{\pi h}{H}\left(i+\frac{1}{2}\right)\right) \cos \left(\frac{\pi w}{W}\left(j+\frac{1}{2}\right)\right) .
	\end{equation}
	Then the 2D DCT for the input image with width $W$ and height $H$ is written by:
	\begin{equation}
		\begin{gathered}
			f_{h, w}^{2 d}=\sum_{i=0}^{H-1} \sum_{j=0}^{W-1} I_{i, j}^{2 d} B_{h, w}^{i, j} \\
			\text { s.t. } h \in\{0,1, \cdots, H-1\}, w \in\{0,1, \cdots, W-1\},
		\end{gathered}
	\end{equation}
	\begin{figure*}[tbp]
		\centering
		\includegraphics[width=0.8\linewidth]{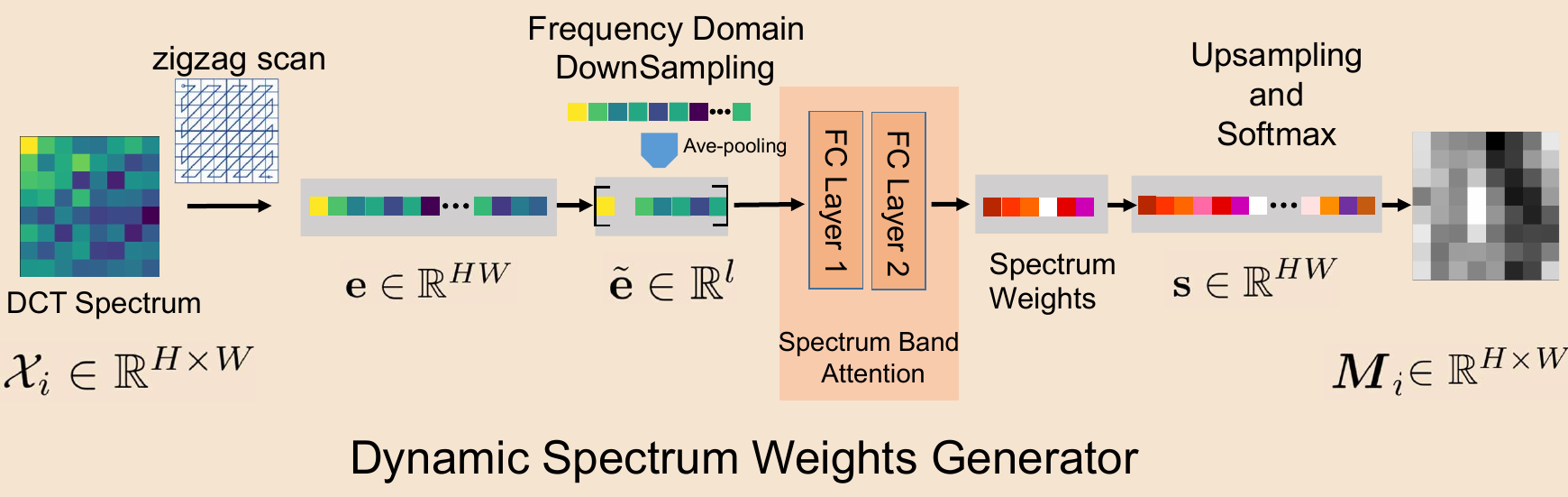}
		\caption{The diagram of the proposed Dynamic Spectrum Weights Generator}
		%\label{fig:intro}
	\end{figure*}
	where $f^{2 d} \in \mathbb{R}^{H \times W}$ is the 2D DCT frequency spectrum, $I^{2 d} \in \mathbb{R}^{H \times W}$ is the input image, Accordingly, the inverse 2D DCT for the image is fomulated as:
	\begin{equation}
		\begin{gathered}
			I_{i, j}^{2 d}=\sum_{h=0}^{H-1} \sum_{w=0}^{W-1} f_{h, w}^{2 d} B_{h, w}^{i, j} \\
			\text { s.t. } i \in\{0,1, \cdots, H-1\}, j \in\{0,1, \cdots, W-1\} .
		\end{gathered}
	\end{equation}
	\par The fast algorithm is \cite{makhoul1980fast} can be utilized to reduce the complexity of 1D-DCT from $\mathcal{O}\left(N^2\right)$ to $\mathcal{O}(N \log N)$. Due to the real-valued structure of the transform, the DCT and Inverse-DCT (IDCT) both have quick $\mathcal{O}(N \log N)$ algorithms that are quicker than FFT. Convolutions may be applied in the DCT domain using a modified form of the Fourier Convolution Theorem \cite{shen1998dct}.
	\subsection{Overall architecture}
	The overall architecture is constructed by
	stacking multiple DSM blocks, as shown in Fig. 1 (a). The basic element of DSM consists of 1) a forward DCT layer, which projects each token into the frequency domain; 2) a Dynamic Spectrum Weights Generator, which adaptively generates the global aggregating weights to modulated frequency features across spectrum bands and channels 3) an inverse DCT layer, which projects the frequency domain feature back to the spatial domain. The key module of the proposed architecture will be introduced in the following sector.
	\par We first divide the input image into $H \times W$ non-overlapping patches and project the flattened patches into $H W$ tokens with dimension $D$. Then
	perform whole image 2D DCT along the spatial dimensions to convert $\boldsymbol{x}$ to the frequency domain:
	\begin{equation}
		\mathcal{X}=\mathcal{DCT}[\boldsymbol{X}] \in \mathbb{R}^{H \times W \times D},
	\end{equation}
	where $\mathcal{DCT}[\cdot]$ denotes the $2 \mathrm{D}$ DCT,  $\mathcal{X}$ is the spectrum of $\boldsymbol{X}$. We can then modulate the spectrum by multiplying a dynamic spectrum aggregator $\boldsymbol{M}$ to the $\mathcal{X}$ :
	\begin{equation}
		\tilde{\mathcal{X}}=\mathcal{X} \odot \boldsymbol{M},
	\end{equation}
	where $\odot$ is the element-wise multiplication operator.
	\par Finally, we project all the modulated frequency tensors back to the spatial domain by reverse DCT:
	\begin{equation}
		\hat{\boldsymbol{X}}=\mathcal{IDCT}[\tilde{\mathcal{X}}].
	\end{equation}
	\subsection{Dynamic Spectrum Weights Generator (DSWG)}
	According to the property of DCT, a single output element from DCT has a component of each of the input tokens in the spatial domain. To this end, the goal of our DSWG module is to generate dynamic spectrum weights $\boldsymbol{M}$ to modulate the transformed DCT frequency bands, which could facilitate the information intersection process reverses back to the spatial domain tokens. The process of dynamic spectrum weights generation is shown in Fig.2. Regarding the frequency bands tensor $\mathcal{X} \in \mathbb{R}^{W \times H \times C}$, we first equally split $\mathcal{X}$ into $C$ parts $\left\{\mathcal{X}_i\right\}_{i=1}^C$ along the channel dimension. Because the spectrum bands of DCT have clear physical meaning and which arranged from left to right and top to bottom in a strictly increasing order of frequencies. To this end, we can flat all the spectrum in $\mathcal{X}_i \in \mathbb{R}^{H \times W }$ in to $\mathbf{e}\in \mathbb{R}^{HW }$ by using zigzag scan, and obtain the one- dimensional embedding. 
	
	\par Intuitively, we can employ multiple fully-connected layers to capture the full spectrum band interaction information. However, the computation complexity will be raised significantly to $\mathcal{O}(H^2 W^2 )$. To solve this problem, We argue that not all information in the frequency range has the perceptual ability. Many DCT-based methods introduce sparsity to DCT blocks through quantization \cite{nash2021generating} or just drop the high-frequency part to reason the global dependency with redundancy elimination \cite{liu2022nommer}. Following this, as shown in Fig.3, we propose to only utilize the average-pooling operation over the inflated spectrum embedding $\mathbf{e}$. We downsampling the frequency band $\mathbf{e} \in \mathbb{R}^{HW}$ to $\tilde{\mathbf{e}} \in \mathbb{R}^{l}$. Then, a two-layer FC design is used to obtain the spectrum band attention weight, which can be formulated as follows:
	\begin{equation}
		\tilde{\mathbf{e}} \in \mathbb{R}^{l}= \mathbf{DownSampling}(\mathbf{zigzag}(\mathcal{X}_i)),
	\end{equation}
	where $Trunc$ and $\mathbf{zigzag}$ are truncation and zigzag scan operation, respectively.
	\begin{equation}
		\mathbf{s} \in \mathbb{R}^{HW} =\textrm{reshape}(\mathbf{W}_2 \sigma\left[\mathbf{W}_1 \operatorname{LayerNorm}(\tilde{\mathbf{e}})\right])
	\end{equation}
	\begin{equation}
		\hat{\mathbf{s}} = \textrm{softmax}(\mathbf{s})
	\end{equation}
	\begin{equation}
		\boldsymbol{M}_i = \mathbf{inverse-zigzag}(\mathbf{UpSampling}(\hat{\mathbf{s}}))
	\end{equation}
	where $\sigma$ is the activation function implemented by Gaussian Error Linear Units (GELU) \cite{hendrycks2016gaussian}, LayerNorm $(\cdot)$ represents the layer normalization \cite{ba2016layer}, which is widely used in MLP-like token mixers.
	$\mathbf{W}_1 \in \mathbb{R}^{K \times l}$ denote the weights of a fully-connected layer, increasing the feature dimension from $l$ to $K $ where $K$ is a fixed value. $\mathbf{W}_2 \in \mathbb{R}^{l \times K }$ refers to the weights of a fully-connected layer reshaping the feature from $K$ back to the original dimension $l$. 
	\par Finally, the $\mathbf{e}$ is further processed by $softmax$ function and followed by inverse zigzag scanning mapping $\mathbb{R}^{HW} \mapsto \mathbb{R}^{H \times W}$ that reshape the embedding $\hat{\mathbf{e}}$ back to get the dynamic spectrum weights $\boldsymbol{M}_i$. To generate a compact model, the mixing weight $\mathbf{W}_1$ and  $\mathbf{W}_2$ are all shared for different channels.

	\subsection{Complexity Analysis}
	Thanks to the fast algorithm, the complexity of the DCT layer is $\mathcal{O}(H W C\left\lceil\log _2(H W)\right\rceil)$, element-wise multiplication takes $\mathcal{O}(HWC)$, compared with standard self-attention $\mathcal{O}\left(H W C^2+H^2 W^2 C\right)$. While the computation cost of DSWG is mainly from the two-layer FCs, with the truncated input  $\mathcal{X} \in \mathbb{R}^{l \times C}$ and $\mathbf{W}_1 \in \mathbb{R}^{K \times l}$, $\mathbf{W}_2 \in \mathbb{R}^{l \times K }$, thus the computation complexity is $\mathcal{O}\left(K l^2 C+K^2 l C\right)$.
	
	\subsection{Architecture Variants}
	Following the hierarchical architecture designs of previous works, e.g., Hire-MLP, \cite{guo2022hire}, we introduce three four-stage variants of DSM architectures by varying the width and depth of the model and number of DSM blocks, denoted as “DSM-S”, “DSM-M” and “DSM-L”, respectively. “DSM-S” version has fewer layers for efficient implementation, while the “DSM-L”
	variant has a larger model size to achieve higher performance. The detailed architecture is provided in the supplemental material.
	%-------------------------------------------------------------------------
	\begin{table}[t]
		\centering
		%\small 
		\footnotesize
		\caption{Comparison of the proposed DSM architecture with existing vision MLP models on ImageNet, the bold numbers are the results of the proposed method} 
		\label{tab-mlp}
		\setlength\tabcolsep{3.4pt}
		\begin{tabular}{l|ccc|c}
			
			\toprule[1.5pt]
			
			\multirow{2}{*}{Model} &  \multirow{2}{*}{Params.} & \multirow{2}{*}{FLOPs} & Throughput &  Top-1 \\ 
			&&&(image / s)&acc. (\%) \\ \hline
			
			EAMLP-14~\cite{guo2022beyond}          & 30M  & -    &771 &78.9 \\
			\hline
			
			Mixer-B/16~\cite{tolstikhin2021mlp}            & 59M  & 12.7G&- & 76.4  \\ 
			\hline
			
			ResMLP-S12~\cite{touvron2022resmlp}                & 15M  & 3.0G  & 1415 &76.6 \\
			ResMLP-S24~\cite{touvron2022resmlp}                 & 30M  & 6.0G  &715& 79.4 \\
			\hline

			gMLP-S~\cite{liu2021pay}                                & 20M  & 4.5G  &-& 79.6 \\
			gMLP-B~\cite{liu2021pay}                                & 73M  & 15.8G &-& 81.6 \\ \hline
			
			S$^2$-MLP-wide~\cite{yu2021s}           & 71M  & 14.0G &-& 80.0 \\
			S$^2$-MLP-deep~\cite{yu2021s}                      & 51M  & 10.5G &-& 80.7 \\ \hline

			ViP-Small/7~\cite{hou2022vision}                  & 25M   & 6.9G  &719& 81.5 \\
			ViP-Large/7~\cite{hou2022vision}                  & 88M   & 24.4G &298& 83.2\\ \hline
			
			AS-MLP-T~\cite{lian2021mlp}              & 28M   & 4.4G  &862& 81.3 \\
			AS-MLP-B~\cite{lian2021mlp}                   & 88M   & 15.2G  &308& 83.3\\  \hline
			
			CycleMLP-B1~\cite{chen2021cyclemlp}    & 15M   & 2.1G  &1040& 78.9 \\
			CycleMLP-B5~\cite{chen2021cyclemlp}         & 76M   & 12.3G &253& 83.2 \\  \hline
			
			$S^{2}$-MLPv2~ \cite{yu2022s2}        & 25M   & 6.9G &-& 82.0 \\
			$S^{2}$-MLPv2~ \cite{yu2022s2}         & 55M   & 16.3G &-& 83.6 \\  \hline
			Hire-MLP-Ti~\cite{guo2022hire}           & 18M   & 2.1G  &-& 79.7 \\
			Hire-MLP-L~\cite{guo2022hire}         & 96M   & 13.4G &-& 83.8 \\  \hline
			ConvMLP-S~\cite{li2021convmlp}       & 9M   & 2.4G  &-& 76.8 \\
			ConvMLP-L~\cite{li2021convmlp}         & 43M   & 9.9G &-& 80.2 \\  \hline
			Wave-MLP-T~\cite{tang2022image}                       & 15M   & 2.1G  &1257& 80.1 \\
			Wave-MLP-B~\cite{tang2022image}                                     & 63M   & 10.2G &341&83.6 \\ \hline
			DSM-S~(ours)                      & 16M   & 2.4G  &1218& \textbf{80.2} \\
			DSM-M~(ours)                         & 30M   & 4.8G  &701& \textbf{82.7} \\	
			DSM-L~(ours)                                   & 90M   & 10.1G  &338 &\textbf{83.8} \\
			\bottomrule[1.5pt]
		\end{tabular}
%		\vspace{-2mm}
	\end{table}
	\begin{table}[t]
		\centering
		%\small 
		\footnotesize
		\caption{Comparison of the proposed DSM architecture with SOTA models on ImageNet. The bold numbers are the results of the proposed method} 
		\label{tab-sota}
		
		\setlength\tabcolsep{2pt}
		\begin{tabular}{l | c |c c c|c}
			
			\toprule[1.5pt]

			\multirow{2}{*}{Model} & \multirow{2}{*}{Family}  & \multirow{2}{*}{Params.} & \multirow{2}{*}{FLOPs} & Throughput &  Top-1 \\ 
			&&&&(image / s)&acc. (\%)\\ \hline

			ResNet18~\cite{he2016deep}                   & CNN    & 12M & 1.8G  &-& 69.8 \\
			ResNet50~\cite{he2016deep}                   &  CNN   & 26M & 4.1G  &- &78.5 \\
			RegNetY-4G~\cite{radosavovic2020designing}   &  CNN   & 21M & 4.0G  &1157& 80.0 \\
			RegNetY-8G~\cite{radosavovic2020designing}   &  CNN  &  39M & 8.0G &592& 81.7 \\
			\hline	
			GFNet-H-S~\cite{rao2021global}               & FFT   &  32M & 4.5G  & -&81.5 \\
			GFNet-H-B~\cite{rao2021global}               & FFT   &  54M & 8.4G  & -&82.9 \\ \hline

			DeiT-S~\cite{touvron2021training}            & Trans  & 22M & 4.6G  & 940&79.8 \\
			PVT-Small~\cite{wang2021pyramid}                 & Trans &  25M & 3.8G  &820& 79.8 \\
			PVT-Large~\cite{wang2021pyramid}                 & Trans &  61M & 9.8G  &367& 81.7 \\
			T2T-ViT-14~\cite{yuan2021tokens} &Trans &22M&5.2G&764&81.5\\
			TNT-S~\cite{han2021transformer} & Trans  & 24M & 5.2G  &428& 81.5 \\
			TNT-B~\cite{han2021transformer}& Trans &  66M & 14.1G  &246& 82.9 \\	
			Swin-T~\cite{liu2021swin}                    & Trans &  29M & 4.5G  &755& 81.3 \\
			Swin-S~\cite{liu2021swin}                    & Trans & 50M & 8.7G  & 437&83.0 \\		
			Swin-B~\cite{liu2021swin}                    & Trans &  88M & 15.4G &278 &83.5 \\

			\hline
			
			DSM-S~(ours)                   & DCT   & 16M   & 2.4G  &1218& \textbf{80.2} \\
			DSM-M~(ours)                       & DCT  & 30M   & 4.8G  &701& \textbf{82.7} \\	
			DSM-L~(ours)                        & DCT           & 90M   & 10.1G  &338 &\textbf{83.8} \\
			\bottomrule[1.5pt]
			
		\end{tabular}
%		\vspace{-3mm}
	\end{table}
	
	\section{Experiments}
	To demonstrate the effectiveness of the proposed DSM architecture, we conduct experiments on a variety of tasks, including image classification, semantic segmentation  ADE20K \cite{ade20k}, and object detection COCO dataset ~\cite{lin2014microsoft}. Ablation investigations are then carried out to confirm the efficiency of each component.
	The experimental results are detailed in the following sections.
	\subsection{ImageNet-1K Classification}
	We evaluate the proposed DSM on the image classification task. The dataset ImageNet \cite{deng2009imagenet} is utilized for benchmarking, which contains
	1.28M training images and over 50k validation images spread over
	1000 classes. To make a fair comparison, we follow the training strategy in DeiT \cite{touvron2021training}. Specifically, we train our model in $4$ NVIDIA A100 GPUs by batch size 1024, and the DSM architecture is implemented by PyTorch. All the “DSM-S”, “DSM-M” and “DSM-L” models are trained for 300 epochs employing AdamW~\cite{loshchilov2017decoupled} optimizer, we set the initial learning rate as $2\times10^{-3}$ and declines to $1\times10^{-6}$ with a cosine decay strategy, and the weight decay is set to 0.05. The warming-up trick for the first 10 epochs is also utilized in our experiments.
	Moreover, We employ the data augmentation strategies following \cite{touvron2021training,hou2022vision}. The length of shrunk spectrum bands is set to be 16 for all the experiments.
	\par \textbf{Comparison with the existing MLP-like architectures.}
	We compare our DSM with multiple MLP-like architectures, e.g., MLP-mixer \cite{tolstikhin2021mlp}, gMLP \cite{liu2021pay}, CycleMLP \cite{chen2021cyclemlp}, AS-MLP \cite{lian2021mlp}, $S^{2}$-MLP and $S^{2}$-MLPv2 \cite{yu2022s2,yu2021s}, Hire-MLP \cite{guo2022hire}. The results are listed in Table 1. We can see that, to reduce the computation complexity, the single scale-based approaches MLP-Mixer,
	ResMLP, gMLP, S2-MLP achieve inferior performance than the hierarchically-designed architecture, such as Hire-MLP and CycleMLP. In contrast, our method achieves a better trade-off between
	the computational cost and performance over the competitors. For instance, with the large-size model DSM-L, compared with previous SOTA Hire-MLP-L~\cite{guo2022hire} (83.8\% accuracy, 13.4GFLOPs), our model also achieves the same accuracy (83.8\%) with significantly similar parameters (90M vs. 96M) and FLOPs (10.1G vs. 13.4G). 
	For the small-size models, our extremely tiny architecture DSM-S achieves 80.2\% with only 16M
	parameters and 2.4G FLOPs, which constantly defeat the Hire-MLP-Ti \cite{guo2022hire}, Wave-MLP-T \cite{tang2022image}, and CycleMLP-B1 \cite{chen2021cyclemlp}. To this end, the effectiveness of the proposed spectrum mixing architecture is consolidated. 
	\par \textbf{Comparisons with CNNs and vision Transformers.}
	To further verify the performance of DSM, We compare the proposed DSM with major CNN and
	transformer-based architectures, the experimental results are shown in Table 2. We can see that our proposed DSM still constantly achieves the best accuracy compared to models with similar FLOPs. Specifically, compared with Swin Transformer \cite{liu2021swin}, which is the state-of-the-art transformer architecture, our DSM obtains higher performance with fewer parameters and less computational complexity. For instance, our DMS-M, with 4.8G FLOPs,
	achieves 82.7\% top-1 accuracy, which significantly surpasses the Swin-T (4.5G FLOPs) with 81.3\% accuracy. This result consolidates that DSM architecture could empower the spectrum token aggregating process efficiently and achieve better performance. According to the above comparison results, we can see that DSM is a powerful architecture for visual recognition tasks.
	\par \textbf{Compared with FFT-based architecture GFNet} The frequency domain token mixing architecture (GFNet-H-B) \cite{rao2021global}, which using static weights to argument all the spectrum, regardless of the input, we achieve $0.9\%$ higher accuracy in Top-1 $(82.9 \%$ vs. $83.8 \%)$. It is verified that equipping with the dynamic spectrum mixing strategy can well capture
	the relationship between spectrum bands derived from input tokens than the static global filter.
	%%%%%%%

	%%%%%%%
	
	\subsection{Semantic Segmentation on ADE20K}
	\noindent\textbf{Settings.} To further verify the performance of our DSM, we conduct the semantic segmentation experiment on ADE20K \cite{ade20k}, which contains 25k images from 150 semantic categories; among them, 20k are used for training, 2k for validation, and 3k for testing. To compare the results fairly, we also adhere to the training guidelines used by the earlier versions of the vision Transformers \cite{chen2021cyclemlp} for the Semantic FPN ~\cite{kirillov2019panoptic} and UperNet ~\cite{upernet} frameworks. The models have been trained using ImageNet-1k with input image size 224$\times$224 and then retrained on ADE20K with a resolution of 512$\times$512. For training the Semantic FPN, we utilize the AdamW optimizer with a learning rate $1.25\times10^{-4}$ and weight decay set to be 0.0001. The Semantic FPN is trained for 40K iterations with a batch size of 32. 
	
	For the UperNet framework, we train the models for 160K iterations . AdamW optimizer is employed with the learning rate $4.25\times10^{-5}$, weight decay 0.01, and batch size is set to 16. The mIoU is then tested using single-scale and multi-scale (MS), where the scale ranges from 0.5 to 1.75 with a 0.25-interval.
	
	\noindent\textbf{Results.} 
	The output from several models for semantic segmentation is displayed in Table 3. Our DSM consistently performs better than the latest models under various parameters and computational configurations. DSM demonstrates a considerable advantage when compared to transformer-based models like PVT, for example, achieving 4.6 percent higher mIoU. Furthermore, it surpasses the Swin-S model and the CycleMLP-B2 by a large margin.

	Additionally, we employ another widely used framework, UperNet~\cite{upernet}, as a backbone for our DSM validation by following~\cite{liu2021swin}. In comparison to the most advanced Swin Transformer~\cite{liu2021swin}, the result suggested DSM outperforms the state-of-the-art Swin Transformer ~\cite{liu2021swin} in terms of MS mIoU and outperforms Swin-T by $+1.7$ mIoU in terms of SS mIoU.We conclude that aggregating various tokens in the frequency domain can capture more comprehensive information and hence improve the semantic segmentation performance.
	\begin{table*}[ht]
		\centering
		\footnotesize
		\renewcommand{\arraystretch}{0.2}
		\setlength\tabcolsep{1.pt}
		\caption{ADE20K validation set semantic segmentation results. A 2048$\times$512 input size is used to calculate FLOPs, and the results come from GFNet, as indicated by the $^\dagger$ symbol. }
		\begin{tabular}{l|c|c|c|c||l|c|c|c|c|c}
			\toprule
			\multicolumn{5}{c||}{Semantic FPN}  & \multicolumn{6}{c}{UperNet} \\
			\midrule  
			Backbone & Param & FLOPs & FPS & SS mIoU & Backbone & Param & FLOPs & FPS & SS mIoU & MS mIoU \\
			\midrule  
			
			PVT-Small & 28M & 163G & 43.9 & 39.8 & Swin-T & 60M & 945G & 18.5, & 44.5 & 46.1 \\
			CycleMLP-B2 & 31M & 167G & 44.5 & 42.4 & AS-MLP-T & 60M & 937G & 17.7\, & - & 46.5 \\
			Hire-MLP-Small & 37M & 174G & 47.3, & {44.3} & Hire-MLP-Small & 63M & 930G & 19.3\, & {46.1} & {47.1} \\
			DSM-S(\textbf{ours}) & 36M & 169G & 48.2\, & \textbf{44.4} & DSM-S(\textbf{ours}) & 60M & 908G & 21.2\, & \textbf{46.2} & \textbf{47.3} \\
			\midrule
			CycleMLP-B3 & 42M & 229G & 31.0, & 44.5 & ResNet-101 & 86M & 1029G & 20.1 & 43.8 & 44.9 \\
			GFNet-Base & 75M & 261G & -\, & 44.8 & Swin-S & 81M & 1038G & 15.2\, & 47.6 & 49.5 \\
			CycleMLP-B4 & 56M & 296G & 23.6, & 45.1 & AS-MLP-S & 81M & 1024G & 14.4, & - & 49.2 \\
			Hire-MLP-Base & 62M & 255G & 31.8\, & {46.2} & Hire-MLP-Base & 88M & 1011G & 16.0\, & {48.3} & {49.6} \\
			DSM-M(\textbf{ours}) & 66M & 258G & 30.3\, & \textbf{46.4} & DSM-M(\textbf{ours}) & 90M & 1100G & 15.1\, & \textbf{48.4} & \textbf{49.7} \\
			\midrule
			
			Swin-B & 53M & 274G & 23.4, & 45.2\rlap{$^\dagger$} & Swin-B & 121M & 1188G & 13.3, & 48.1 & 49.7 \\
			CycleMLP-B5 & 79M & 343G & 22.9, & 45.6 & AS-MLP-B & 121M & 1166G & 11.0, & - & 49.5 \\
			Hire-MLP-Large & 99M & 366G & 24.5\, & {46.6} & Hire-MLP-Large & 127M & 1125G & 13.7\, & {48.8} & {49.9} \\
			DSM-L(\textbf{ours}) & 93M & 349G & 24.9\, & \textbf{46.8} & DSM-L(\textbf{ours}) & 119M & 1109G & 13.9\, & \textbf{48.9} & \textbf{49.9} \\
			\bottomrule
		\end{tabular}
%		\vspace{-0.3cm}
		
%		\vspace{-0.2cm}
		\label{table:seg}
	\end{table*}
	
	%\vspace{-8mm}
	\subsection{Object Detection on COCO}
	\noindent\textbf{Settings.} We perform the object detection and instance segmentation tests using the COCO 2017 benchmark ~\cite{lin2014microsoft}, which includes 118K training pictures and 5K validation images. Two popular detectors, RetinaNet ~\cite{lin2017focal} and Mask R-CNN ~\cite{he2017mask}, both employ DSM as their backbone. We train the model using the AdamW ~\cite{loshchilov2017decoupled} optimizer for 12 epochs following the training recipe in \cite{wang2021pyramid} to ensure a fair comparison. The resized images are used to train models, with the longest side being no more than 1333 pixels and the shorter side ranging from 480 to 800 pixels. the batch size is set as 16 with a starting learning rate of 0.0001. While the backbones are initialized with pre-trained weights from ImageNet.
	
	\noindent\textbf{Results.} In Table 4, we provide the object identification and instance segmentation results obtained using various training schedules and frameworks. According to Table 4, under comparable FLOPs limitations, when compared to the current models, DSM-based RetinaNet and Mask R-CNN, consistently outperform CNN-based ResNet, transformer-based PVT, and MLP-based CycleMLP. Using RetinaNet as the backbone, our DSM-S consistently outperforms ResNet50 and CycleMLP-B2, bringing +7.2 AP gains over ResNets and +2.6 AP gains over CycleMLP-B2 with significantly greater model size and FLOPs. As demonstrated in Table 4, DSM-M-based Cascade Mask R-CNN outperforms Wave-MLP-M in both box AP and mask AP with fewer FLOPs. Our DSM achieves much greater performance (45.4 box AP and 41.1 mask AP) with fewer parameters (58.6M) and lower computational cost (252.1G) than Swin-S, which has 44.8 box AP and 40.9 mask AP with 69.1M parameters and 353.8G FLOPs. According to the findings, DSM may make a fantastic foundation for object identification. 
	
	%When compared to the current models, the suggested Wave-MLP clearly outperforms %them for both RetinaNet and Mask R-CNN. For instance, Wave-MLP-T obtains 40.4\% AP %with RetinaNet 1$times$ using just 25.3M parameters and 196.3G FlOPs, outperforming %CycleMLP-B1 (38.6 AP) with a similar model size by 1.8 AP. The performance gains %are equally substantial when Mask R-CNN is used as the detector. Our Wave-MLP-S %achieves much greater performance (44.0 box AP and 40.0 mask AP) with fewer %parameters (47.0M) and lower computational cost (250.3G) than Swin-T, which has %42.2 box AP and 39.1 mask AP with 47.8M parameters and 264.0G FLOPs.
	\begin{table*}[t]
		\centering
		\scriptsize
		\caption{Results of object detection and instance segmentation on COCO val2017.}
		\label{tab-coco}
		\resizebox{\textwidth}{35mm}{
			\setlength\tabcolsep{1pt}
			
			\begin{tabular}{l|c |lcc|lcc| c|lcc|lcc}
				
				\toprule[1.5pt]
				\multirow{2}{*}{Backbone} &\multicolumn{7}{c|}{RetinaNet 1$\times$} &\multicolumn{7}{c}{Mask R-CNN 1$\times$} \\
				\cline{2-15} 
				& Params. / FLOPs & AP &AP$_{50}$ &AP$_{75}$ &AP$_S$ &AP$_M$ &AP$_L$ & Params. / FLOPs& AP$^{\rm b}$ &AP$_{50}^{\rm b}$ &AP$_{75}^{\rm b}$  &AP$^{\rm m}$ &AP$_{50}^{\rm m}$ &AP$_{75}^{\rm m}$\\ \hline

				ResNet50                & 37.7M / 239.3G & 36.3 & 55.3 & 38.6 & 19.3 & 40.0 & 48.8 & 44.2M / 260.1G & 38.0 & 58.6 & 41.4 & 34.4 & 55.1 & 36.7 \\
				
				Swin-T & 38.5M / 244.8G & 41.5 &62.1&44.2& 25.1 & 44.9 & \textbf{55.5} & 47.8M / 264.0G & 42.2 & 64.6 & 46.2 & 39.1 & 61.6 & 42.0 \\ 
				PVT-Small          & 34.2M /226.5G & 40.4 & 61.3 & 43.0 & 25.0 & 42.9 & 55.7 & 44.1M / 245.1G & 40.4 & 62.9 & 43.8 & 37.8 & 60.1 & 40.3 \\
				CycleMLP-B2                           & 36.5M / 230.9G& 40.9 & 61.8 & 43.4 & 23.4 & 44.7 & 53.4 & 46.5M /249.5G & 41.7 & 63.6 & 45.8 & 38.2 & 60.4 & 41.0 \\
				
				Wave-MLP-S                           & 37.1M / 231.3G &{43.4} &64.4&46.5&\textbf{26.6}&47.1&57.1 & 47.0M /250.3G &{44.0} &\textbf{65.8}&48.2&40.0&63.1&\textbf{42.9} \\
				DSM-S(\textbf{ours})                           & 35.1M / 219.2G &\textbf{43.5} &\textbf{65.5}&\textbf{46.6}&\textbf{26.6}&\textbf{47.2}&\textbf{57.3} & 46.2M /248.1G &\textbf{44.1} &65.6&\textbf{48.4}&\textbf{40.1}&\textbf{63.2}&\textbf{42.9} \\
				\hline
				
				ResNet101              & 56.7M / 315.4G & 38.5 & 57.8 & 41.2 & 21.4 & 42.6 & 51.1 & 63.2M / 336.4G & 40.4 & 61.1 & 44.2 & 36.4 & 57.7 & 38.8 \\

				Swin-S&59.8M / 334.8G & 44.5 &65.7&47.5& 27.4 & 48.0 & 59.9 & 69.1M / 353.8G & 44.8 & 66.6 & 48.9 & 40.9 & 63.4 & 44.2 \\
				PVT-Medium         & 53.9M / 283.1G & 41.9 & 63.1 & 44.3 & 25.0 & 44.9 & 57.6 & 63.9M / 301.7G & 42.0 & 64.4 & 45.6 & 39.0 & 61.6 & 42.1 \\
				
				CycleMLP-B3                           & 48.1M / 291.3G & 42.5 & 63.2 & 45.3 & 25.2 & 45.5 & 56.2 & 58.0M / 309.9G & 43.4 & 65.0 & 47.7 & 39.5 & 62.0 & 42.4 \\ 
				
				Wave-MLP-M                           & 49.4M / 291.3G &{44.8} &65.8&47.8&\textbf{28.0}&48.2&59.1&59.6M / 311.5G&{45.3}&67.0&\textbf{49.5}&41.0&64.1&44.1 \\ 
				DSM-M(\textbf{ours})                 & 48.1M / 288.1G &\textbf{44.9} &\textbf{65.9}&\textbf{47.9}&26.6&\textbf{48.3}&\textbf{59.4} & 58.6M /252.1G &\textbf{45.4} &\textbf{67.2}&49.4&\textbf{41.1}&\textbf{64.2}&\textbf{44.3} \\
				
				\hline
				
				PVT-Large          & 71.1M / 345.7G & 42.6 & 63.7 & 45.4 & 25.8 & 46.0 & 58.4 & 81.0M / 364.3G& 42.9 & 65.0 & 46.6 & 39.5 & 61.9 & 42.5 \\
				CycleMLP-B4                           & 61.5M / 356.6G & 43.2 & 63.9 & 46.2 & 26.6 & 46.5 & 57.4 & 71.5M / 375.2G & 44.1 & 65.7 & 48.1 & 40.2 & 62.7 & 43.5 \\
				CycleMLP-B5                           & 85.9M / 402.2G & 42.7 & 63.3 & 45.3 & 24.1 & 46.3 & 57.4 & 95.3M / 421.1G & 44.1 & 65.5 & 48.4 & 40.1 & 62.8 & 43.0 \\
				
				Wave-MLP-B &66.1M / 333.9G&\textbf{44.2}&65.1&47.1&27.1&47.8&58.9&75.1M / 353.2G&\textbf{45.7}&67.5&50.1&27.8&{49.2}&\textbf{59.7}\\ 
				
				DSM-L(\textbf{ours})  & 64.1M / 300.1G &{44.1} &\textbf{65.4}&\textbf{47.3}&{27.4}&{47.9}&\textbf{59.0} & 71.0M /348.3G &\textbf{45.7} &\textbf{67.7}&\textbf{50.3}&\textbf{41.7}&\textbf{64.7}&{45.4} \\
				\bottomrule[1.5pt]
		\end{tabular}}
		
%		\vspace{-4mm}
	\end{table*}
	\section{Ablation Studies}
	\begin{minipage}{\textwidth}
		
		\begin{minipage}[htb]{0.45\textwidth}
			\makeatletter\def\@captype{table}
			\footnotesize
			\caption{Ablation Study on Dynamic Spectrum Weight.}
			\begin{tabular}{l|c|c}
				\hline Model & generating methods  & Top-1 (\%) \\
				\hline DSM-S & Random Weight  & $70.5$ \\
				DSM-S & All-pass Filter  & $76.4$ \\
				DSM-S & DSM-Dynamic  & $80.2$ \\
				\hline
			\end{tabular}
			
			\label{sample-table}
		\end{minipage}
		\begin{minipage}[htb]{0.40\textwidth}
			\makeatletter\def\@captype{table}
			\footnotesize
			\caption{Albantion Study on Spectrum Length for DSM-S model.}
			\begin{tabular}{l|c|c|c}
				\toprule[1.1pt]
				Length& Params. & FLOPs   &  Top-1 accuracy (\%) \\ \midrule 
				
				$4\times4$         &12M& 1.8G  & 72.7 \\
				$8\times8$         &15M& 2.1G  & 76.9 \\
				$10\times10$ &16M &2.4G& 80.2  \\ 
				$12\times12$ &16.4M &3.8G& 80.3  \\
				$16\times16$ &16.8M &6.9G& 80.3  \\
				All &18M &8.3G&80.4  \\ 
				\bottomrule[1.5pt]
			\end{tabular}
			
			\label{sample-table}
		\end{minipage}
	\end{minipage}
	\subsection{The Effectiveness of Dynamic Spectrum Weight}
	To evaluate the effectiveness of the proposed Dynamic Spectrum Weight generator module, we designed three versions of experiments. (1) Only using the random values as the spectrum mixing weight. (2) All-pass filter, i.e., set all the weights to 1, left the spectrum untouched. In other words, merely utilize original 2D DCT to mix all the tokens, a similar way to FNet \cite{eckstein2022fnet}. (3) Our Dynamic Spectrum Weight generator. The comparison results are listed in Table 5. We can see that the random weight leads to significant performance degradation (70.5\% Top-1), while the all-pass filter achieves much better accuracy than the random one. Among all the weight generation mechanisms, our DSM demonstrates the best performance. This result verifies that the dynamically generated spectrum weight empowers the token mixer with improved representation ability. 
	\subsection{Influence of the Spectrum Length}
	To evaluate the influence of spectrum length $l$ on the performance, We compare the performance of several settings in Table 6 on ImageNet-1K.
	Intuitively, the longer spectrum length is beneficial to
	fine-grained modeling details of the input tokens and tends to achieve higher recognition accuracy. 
	Surprisingly, as shown in the table, even employing only $8\times8$ spectrum bands can achieve $76.9\%$ accuracy. As the length grows from 32 to 64, the accuracy remains unchanged. This result confirms that compressed DCT spectrums can reduce the computation cost for the proposed DSM without sacrificing accuracy. 
	
	\section{Conclusion}
	We propose a dynamic token mixing architecture in the frequency domain for the visual recognition task, representing each token as a DCT spectrum. With the dynamically generated weight, all the tokens are aggregated with varying contents from different input images.
	Extensive experiments show that the proposed DSM outperforms the existing MLP-like architectures and can also be used as a strong backbone for the image classification task. In the future, we will further explore the DSM architecture for downstream dense prediction tasks, such as object detection and semantic segmentation.
	%%%%%%%%% REFERENCES
	{\small
		\bibliography{egbib}
	}
\end{document}